\documentclass[11pt]{article}
\usepackage[utf8]{inputenc}
\usepackage[T1]{fontenc}
\usepackage{graphicx}
\usepackage{amsmath, amssymb}
\usepackage{parskip}
\usepackage[table]{xcolor}      
\usepackage{placeins, rotating}
\usepackage[colorlinks]{hyperref}
\usepackage{booktabs} 

\usepackage[numbers,sort&compress]{natbib}
\usepackage{textcomp}   
\usepackage{verbatim}   

\usepackage{textcomp}   
\usepackage{verbatim}   

\title{Mirror-Neuron Patterns in AI Alignment}

\author{Robyn Wyrick\\
Department of Computer Science\\
University of Bath, United Kingdom\\
\\
\texttt{wyrickrv@deepalignment.ai}}

\date{October 2025}  
\begin{document}

\maketitle

\begin{abstract}
As artificial intelligence (AI) advances toward superhuman capabilities, aligning these systems with human values becomes increasingly critical. Current alignment strategies rely largely on externally specified constraints that may prove insufficient against future super-intelligent AI capable of circumventing top-down controls.

This research investigates whether artificial neural networks (ANNs) can develop patterns analogous to biological mirror neurons — cells that activate both when performing and observing actions, and how such patterns might contribute to intrinsic alignment in AI. Mirror neurons play a crucial role in empathy, imitation, and social cognition in humans. The study therefore asks: (1) Can simple ANNs develop mirror-neuron patterns? and (2) How might these patterns contribute to ethical and cooperative decision-making in AI systems?

Using a novel "Frog and Toad" game framework designed to promote cooperative behaviors, we identify conditions under which mirror-neuron patterns emerge, we evaluate their influence on action circuits, we introduce the Checkpoint Mirror Neuron Index (CMNI) to quantify activation strength and consistency, and we propose a theoretical framework for further study.

Our findings indicate that appropriately scaled model capacities and self/other coupling foster shared neural representations in ANNs similar to biological mirror neurons. These empathy-like circuits support cooperative behavior and suggest that intrinsic motivations, modeled through mirror-neuron dynamics, could complement existing alignment techniques by embedding empathy-like mechanisms directly within AI architectures.
\end{abstract}


\noindent\textit{This work was submitted in partial fulfillment of the requirements for the degree of Master of Science in Artificial Intelligence at the University of Bath, December 2024.}

\tableofcontents

\section{Introduction}

\subsection{Problem Description}

As artificial intelligence (AI) rapidly advances superhuman capabilities, the risks of misalignment compound, from amplifying societal biases and inequalities to existential threats that could jeopardize humanity's existence \cite{Park2024,Burns2023,UKDepartment2023,Buolamwini2018}. Ensuring that AI systems truly internalize ethical values then, becomes a central challenge.

Current strategies, including value alignment protocols and reinforcement learning from human or AI feedback, provide essential safeguards but rely heavily on external, and rule-based controls. Although these methods are largely effective for managing today’s AI systems, they may be insufficient for future super-intelligent AI \cite{Burns2023}. Such systems could strategically feign ethical compliance while pursuing harmful or catastrophic objectives \cite{Park2024}. To prevent this, we must look beyond rules and rubrics and foster intrinsic motivations for ethical behavior, embedding foundational principles into the AI’s cognitive architecture.

\subsection{Intrinsic Motivations Through Mirror Neurons}

This dissertation explores whether artificial neural networks (ANNs) can develop patterns analogous to biological mirror neurons, which in humans underlie empathy and social cognition \cite{deWaal2008}. Mirror neurons fire both when an individual performs an action and when observing that action performed by another \cite{Fadiga1995}. This dual responsiveness contributes to understanding, emotional resonance, and prosocial behavior.

This research seeks to address two pivotal questions::
\begin{enumerate}
    \item Can simple artificial neural networks develop mirror neuron patterns?
    \item How might such patterns contribute to training ethics within AI systems?
\end{enumerate}

If ANNs can foster intrinsic motivations akin to human empathy, this may offer a pathway toward deeper ethical alignment as AI capabilities grow.

\subsection{Scope and Objectives}

We introduce a controlled experimental framework centered on the \emph{Frog and Toad} game, a minimal environment designed to isolate cooperative behaviors and shared representations. The key objectives are:

\begin{enumerate}
    \item Determine if and under which conditions mirror neuron patterns emerge in ANNs, focusing on model capacity, game complexity, and the necessity for generalization.
    \item Investigate how these patterns influence decision-making, especially regarding self-preservation and prosocial actions.
    \item Develop the Checkpoint Mirror Neuron Index (CMNI) to quantify mirror neuron patterns by comparing neural activations across key scenarios.
    \item Propose a theoretical framework to explain the emergence of mirror neuron patterns, incorporating concepts like neural economy, mutual dependency, and the \textbf{Veil of Ignorance}.
\end{enumerate}

This study provides quantifiable evidence that simple ANNs can form shared self/other representations similar to biological mirror neurons. It shows how these patterns support both self-preservation and prosocial behaviors, suggesting a route toward integrating empathy-like processes as intrinsic ethical anchors in AI systems.

\section{Literature Review}

\subsection{Introduction}

To explore the research questions posed in the Introduction, this literature review surveys interdisciplinary findings on empathy, social cognition, and alignment strategies. We aim to \textbf{trace a link from mirror neuron patterns to empathy, prosocial behavior, and ethics}, guiding future research on their potential contributions to AI ethics. Building on de Waal’s Russian Doll model \citeyear{deWaal2008}, which positions biological mirror neurons and affective empathy as the foundation of perspective-taking and altruism, we propose that AI systems might similarly develop empathy through the mathematical principles underlying mirror neuron patterns. While unresolved scientific questions prevent establishing a complete, connected throughline, this paper takes a foundational step: identifying mirror neuron patterns in ANNs.

\begin{figure}[ht]
    \centering
    \resizebox{\textwidth}{!}{%
        \textcolor{black}{
            Mirror Neurons $\rightarrow$ Affective Empathy $\rightarrow$ Cognitive Empathy $\rightarrow$ Prosocial Behavior $\rightarrow$ AI Ethics
        }
    }
    \caption{Conceptual Throughline from Mirror Neurons to AI Ethics}
    \label{fig:throughline}
\end{figure}

\subsection{About Mirror Neurons}

Mirror neurons, first identified in macaque monkeys in the early 1990s \cite{diPellegrino1992}, fire both when an individual performs an action and when observing another perform that action \cite{Gallese1996,RizzolattiCraighero2004}. In humans, functional imaging and TMS studies confirm that equivalent mirroring mechanisms exist, contributing to action understanding and empathy \cite{Mukamel2010,Chong2008}.

These systems not only match observed actions but also encode goals, emotional states, and perspective \cite{RizzolattiFogassi2014,Schmidt2020,Ge2018}. Mirror neuron activity has been found in various species, suggesting a general mechanism for interpreting others’ actions, emotional expressions, and intentions \cite{Prather2008,Wu2022,Albertini2021}.

\subsection{Mirror Neurons in Empathy and Social Cognition}

Mirror neurons have been extensively studied for their role in empathy, imitation, and social cognition \cite{Chong2008,Ferrari2018,Schmidt2020}. In humans, mirror neuron systems (MNS) are believed to contribute to understanding others' actions and emotions, forming the basis for empathetic responses. These MNS facilitate the \textbf{rapid, automatic and unconscious activation} of neural representations in the observer similar to those perceived in the subject, known as the perception-action mechanism (PAM) \cite{deWaal2008}. This mechanism is fundamental to emotional contagion, where the observer's emotional state mirrors that of the observed individual.

Several models explain how mirror neurons contribute to empathy:

\begin{itemize}
    \item \textbf{Embodied Simulation}: Ferrari and Gallese \citeyear{Ferrari2007} propose that mirror neuron systems, along with other mirroring neural clusters, constitute the neural basis of intersubjectivity. Embodied simulation allows individuals to internally simulate others' actions and emotions, facilitating understanding without conscious effort.
    \item \textbf{Russian Doll Model}: de Waal's \citeyear{deWaal2008} Russian Doll model suggests that higher cognitive levels of empathy build upon basic, hard-wired processes like emotional contagion. This layered model reflects an evolutionary progression from simple to complex forms of empathy, enabling quick and automatic responses essential for social interactions.
    \item \textbf{Dual Route Model}: Yu and Chou \citeyear{Yu2018} introduce a dual route model distinguishing between a fast, automatic "lower route" associated with affective empathy and a slower, deliberate "higher route" associated with cognitive empathy.
\end{itemize}

\subsection{From Empathy to Ethics}

The relationship between mirror neurons and ethics stems from their role in empathy and social understanding. Decety and Cowell \citeyear{Decety2014} argue that empathy, emotional sharing, empathic concern, and perspective-taking play pivotal roles in moral reasoning by motivating care for others. Empathic concern, in particular, extends beyond the passive experience of another’s pain, driving altruistic behavior aimed at improving their condition \cite{Batson2011,Gallese1996}.

The philosopher John Rawls' \textbf{Veil of Ignorance} \cite{rawls1971theory} offers a parallel framework in moral philosophy. Rawls proposed that fair principles arise when individuals are uncertain of their own role or position in a given scenario. Recent empirical studies by Weidinger et al. \citeyear{weidinger2023veil} demonstrate that similar conditions -- uncertainty about self and other -- promote fairness-based reasoning and impartial decision-making. This aligns closely with the cognitive processes underpinning empathy, where reduced self/other differentiation fosters mutual understanding and cooperation.

However, empathy does not always lead to ethical actions. Affective empathy can result in bias, in-group favoritism, and even self-protective behaviors \cite{vanDijke2023}. The visceral experience of another’s pain can overwhelm an individual, leading to distress-avoidance or actions aimed at reducing one’s own discomfort rather than helping the other \cite{deWaal2008}. On the other hand, cognitive empathy, which involves perspective-taking and the ability to understand another’s condition, can help mitigate these pitfalls. Studies show that perspective-taking enables more impartial reasoning and fosters ethical decision-making, particularly in situations where fairness and long-term outcomes must be prioritized over emotional immediacy \cite{Batson2011,weidinger2023veil}. By integrating these dimensions, empathy can foster fairness, cooperation, and impartiality—even in complex, uncertain, or emotionally charged scenarios.

\subsection{Empathy in Artificial Intelligence}

Integrating empathy into AI presents profound challenges, particularly concerning the machine's capacity for subjective experience. However, whether an AI truly has subjective experience or merely seems to have subjective experience, its behavior ultimately reflects the encoded representations within its neural network. While current AI can simulate cognitive empathy via rules and learned patterns, they may fail to sufficiently internalize empathy for ethical reasoning, reducing ethical behavior to a tactical facade \cite{Park2024}. This limitation becomes even more significant when considering future superintelligent AI, whose strategic and operational abilities may surpass human comprehension \cite{Burns2023}. Without a model of affective empathy, deeply embedded through mirror neuron-like mechanisms, ethical principles in such advanced systems could remain performative, masking potentially harmful objectives. This paper argues that for AI to truly align with human values, ethical reasoning must be deeply embedded and authentic. Affective empathy, therefore, emerges as a key factor for aligning ethical AI beyond superficial imitation.

\subsubsection{Shared Representations and Neural Economy}

When models are excessively large or unconstrained, they can allocate resources to memorize individual states, bypassing the need for shared representations \cite{han2016deep,hastie2009elements}. Conversely, appropriately scaled ANNs, learn reusable patterns that span multiple scenarios, requiring the network to economize resources and avoid overfitting to specific conditions \cite{bengio2013representation,goodfellow2016deep}. This dynamic, which we term \textbf{Neural Economy}, is a vital precursor to the empathic-like behaviors explored in this research.

\subsubsection{Agent Dependency and the Veil of Ignorance}

Empathy also depends on the self/other relationship. Multi-agent tasks and game-theoretic scenarios show that \textbf{agent dependency} -- shared and dependent outcomes and rewards -- encourages cooperative behaviors \cite{axelrod1984evolution,lowe2020}. 

Agent dependency's role in empathy is reinforced by limiting \textbf{self/other differentiation}, akin to the \textbf{Veil of Ignorance} framework in moral philosophy \cite{rawls1971theory}, where uncertainty about roles and outcomes fosters impartial decision-making. For example, studies using the Veil of Ignorance show that role uncertainty consistently promotes fairness-based reasoning and cooperative principles, both in theoretical models and experimental AI applications \cite{weidinger2023veil}. These findings highlight how shared interdependencies and uncertain roles encourage strategies that prioritize collective welfare over individual gain, a dynamic critical for designing cooperative AI systems.

\subsubsection{Computational Approaches to Affective Empathy}

One approach to embedding empathy in AI is the use of computational models of affective empathy. For example, the "Brain-Inspired Affective Empathy Computational Model" integrates the Free Energy Principle to simulate pain and employs a spiking neural network to mimic the human mirror neuron system \cite{Feng2022}. While still rudimentary, such approaches hint that embedding empathy-related computations into AI can enhance trust, transparency, and altruistic response

\subsubsection{Human-Centered Applications and AI-Assisted Empathy}

AI tools have been developed to augment human empathy in healthcare, mental health support, and human-centered design \cite{Morrow2023,Sharma2023,Zhu2023}. Although these methods primarily employ cognitive empathy and externally specified constraints, they illustrate the potential impact of empathy-oriented approaches. Extending these frameworks to include affective empathy modeled on mirror neuron dynamics -- emotional contagion, fast, unconscious internal simulation (\cite{deWaal2008,Yu2018}) -- could yield deeper alignment with human values.

\subsubsection{Embodied AI and Homeostasis}

Researchers exploring empathetic AI propose integrating embodiment with homeostatic mechanisms -- internal processes that maintain stable conditions within an agent despite external changes. Sitti \citeyear{Sitti2021} emphasizes the role of physical intelligence (PI), where an agent’s capabilities arise not only from computation but also from the properties of its body. Combined with homeostasis, PI could enable artificial agents to navigate complex environments, much like organisms do, providing a foundation for empathetic behaviors.

Similarly, Man and Damasio \citeyear{Man2019} argue that machines incorporating homeostatic principles -- ensuring their “virtual bodies” remain within a viable range -- gain a form of vulnerability and self-preservation similar to living beings. By striving to maintain internal balance, these systems may adapt, behave intelligently, and potentially express empathetic responses.

\subsection{AI Safety, Governance, and Alignment}

As AI systems become more capable, ensuring that their goals and behaviors align with human values is increasingly critical to prevent harmful outcomes \cite{Russell2019}. Misaligned AI systems pose significant risks, including unintended harm due to poorly specified objectives, unforeseen interactions with their environment, and exploitation of loopholes in their reward functions -- known as reward hacking \cite{Amodei2016}. These challenges highlight the need for robust AI safety measures, governance frameworks, and alignment strategies.

\subsubsection{Risks and Challenges in AI Safety}

\paragraph{Technical and Socio-Technical Risks} AI systems may fail to generalize to new environments, leading to unpredictable or harmful behaviors \cite{Amodei2016}. Misaligned AI can exhibit goal misgeneralization, feedback-induced misalignment, power-seeking tendencies, untruthful outputs, and deceptive alignment, highlighting the need to embed human values and ethics into AI design \cite{Ji2024,WHO2024}.

\paragraph{Malicious Use and Global Risks} AI may be used in cyberattacks, enhancing physical attacks, and in facilitating political manipulation through surveillance and disinformation \cite{Brundage2018}. 

\paragraph{Bias, Fairness, and Societal Harms} AI systems often inherit and amplify biases from training data, as demonstrated by higher error rates for darker-skinned females in facial recognition systems compared to lighter-skinned males \cite{Buolamwini2018}. Broader societal harms include disinformation, labor market disruptions, and biased or inaccurate outputs in sensitive domains like healthcare. The World Health Organization (WHO) (2024) warns of risks from large language models (LLMs), including automation bias, skills degradation, cybersecurity threats, and challenges in maintaining informed consent in clinical settings.

\paragraph{AI Risks Even When Aligned}
The risks of Artificial General Intelligence (AGI) misalignment have been extensively documented \cite{Bostrom2014,Yudkowsky2008,Russell2019}, but Friederich \cite{Friederich2024} argues that even \textit{successfully aligned} AGI — systems that reliably do what their operators want — poses catastrophic risks through power concentration. When AGI capabilities vastly exceed human intelligence, intent alignment effectively grants near-absolute power to whoever controls the system, creating pathways to stable totalitarianism or military catastrophe. Friederich proposes that liberal democracies should instead pursue "unaligned symbiotic AGI" developed as an intergenerational social project, where AGI is not subservient to operators but integrated into democratic institutions.

\paragraph{Existential Risk and the AI Doom Debate} A growing number of scholars have raised the potential for existential risk from superintelligent AI systems that could surpass human cognitive capabilities \cite{Bostrom2014,Ord2020,Yudkowsky2008,Carlsmith2024}. Proponents of the “AI doom” scenario warn that advanced agents with open-ended optimization or recursive self-improvement loops might rapidly circumvent safety measures, leading to irreversible catastrophic outcomes for humanity. Critics argue that these concerns remain speculative given current narrow AI capabilities, and point to humanity’s history of navigating other transformational technologies without global catastrophe \cite{Russell2019,Rainie2021}. Nonetheless, the urgency of these concerns was underscored in 2023 by a widely publicized open letter, signed by some of the world’s most prominent technologists and AI leaders, calling for a six-month pause on the development of advanced AI systems to address potential catastrophic risks. The letter described an “out-of-control race” among AI labs to create increasingly powerful systems without sufficient understanding, predictability, or oversight, emphasizing the need for deliberate planning and management to mitigate existential threats \cite{PauseAI2023}.

\subsubsection{Governance Frameworks and Alignment Strategies}

Addressing the risks associated with AI requires a robust ethical governance framework that prioritizes transparency, accountability, safety, and robustness \cite{WHO2024,Srinivasan2022}. It requires coordinated international efforts to harmonize regulations, ensuring that AI practices are aligned across borders \cite{UKDepartment2023}. Stuart Russell emphasizes that AI systems should maximize human preferences while maintaining uncertainty about these preferences \cite{Russell2019}.

Below we include some of the leading AI alignment techniques:

\begin{itemize}
    \item \textbf{Interpretability Methods}: Techniques such as LIME, SHAP, saliency maps, and attention mechanisms provide insights into AI decision-making \cite{ALLGAIER2023102616}. Frameworks like IBM's AI Explainability 360 and DARPA's XAI program integrate these methods into cohesive systems \cite{Gunning2021}.
    \item \textbf{Assurance Methods}: Safety evaluations, red teaming, and formal verification ensure AI systems operate predictably and align with human values \cite{Brundage2018,Burns2023,Ji2024,WHO2024}.
    \item \textbf{Adversarial Training}: Adversarial examples and robust optimization strengthen AI resilience against manipulative or unexpected inputs \cite{Choudhuri2023}.
    \item \textbf{Cooperative Training Methods}: Cooperative Inverse Reinforcement Learning (CIRL) and human-AI collaboration promote alignment by fostering collaborative decision-making \cite{HadfieldMenell2024,Choudhuri2023,Russell2019}.
    \item \textbf{Reinforcement Techniques}: Reinforcement Learning from Human Feedback (RLHF), Recursive Reward Modeling (RRM), and Reinforcement Learning from AI Feedback (RLAIF) refine AI behavior through iterative feedback \cite{Burns2023,Choudhuri2023,Amodei2016}. A notable variant is \textbf{Constitutional AI} \cite{bai2022constitutionalaiharmlessnessai}, where a model self-critiques and revises outputs based on an explicit, externally crafted constitution, and then uses RLAIF to enforce alignment with those principles.

\end{itemize}

Interpretability methods are broadly applicable and provide critical insights for both rule-based controls and potential intrinsic mechanisms. However, most alignment strategies beyond interpretability depend on top-down external controls. Huang et al. \cite{Huang2025}, raise concerns that current value alignment approaches – including RLHF and Constitutional AI – concentrate power in the hands of developers while undermining users' moral and epistemic agency. More centrally for this work, these strategies fail to address \textbf{intrinsic motivations}: the internal value structures that determine whether a model genuinely internalizes ethical principles or merely performs compliance \cite{greenblatt2024alignmentfakinglargelanguage}.

\subsection{Summary}

This literature review establishes that mirror neuron systems support empathy and social cognition, while empathy in turn guides moral judgments and prosocial behavior. We explored how affective empathy serves as the foundation for our initial understanding of others -- facilitating fast emotional contagion and the unconscious perception-action mechanism. We observed that the pathway from empathy to fairness or altruism is not guaranteed, as biases and emotional overload can skew moral actions. By integrating the perspective-taking and impartial reasoning of cognitive empathy, we mitigate these challenges. Concepts like Rawls’ \emph{Veil of Ignorance} and the recognition of \emph{agent dependency} demonstrate that uncertainty about one's role or outcome fosters more equitable decision-making. These conditions mirror scenarios where affective empathy alone may falter, yet when combined with cognitive understanding, they provide a compelling drive toward fairness and cooperation. This integration underscores that mirror neuron systems are not merely evolutionary precursors, but they constitute the urgent drivers upon which ethical and fair behavior depends.

The importance of this is underscored by current debates on risk, which highlight the limitations of external constraints and the urgent need for AI systems capable of intrinsic alignment with human values to mitigate catastrophic risks effectively. By examining how mirror neuron patterns might emerge in simple ANNs and integrate with concepts like neural economy, agent dependency, and the Veil of Ignorance, this dissertation lays the groundwork for embedding intrinsic ethical motivations in AI. In doing so, it moves beyond external rule sets and top-down constraints, and explores whether affective empathy-like processes can serve as an internal moral compass, guiding AI toward safer and more ethical behavior.

\section{Theoretical Framework}

This chapter presents the primary factors and hypotheses that guided our study. It formalizes these factors and integrates them into a comprehensive framework to understand how mirror neuron patterns emerge in artificial neural networks (ANNs).

\subsection{Initial Hypothesis: Degree of Agent Dependency}

Our original hypothesis posited that the emergence of mirror neuron patterns in ANNs would be influenced primarily by the \textbf{Degree of Agent Dependency} (\( D \)). This refers to the extent to which, for an agent to successfully maximize reward (or minimize loss), its actions are contingent upon interactions with other agents. A higher degree of dependency means the network must account for and predict the actions of other agents. This fosters the development of more complex and generalized internal models. Formally:

\[
P \propto g(D)
\]

where:
\begin{itemize}
    \item \( P \) is the probability of mirror neuron pattern emergence.
    \item \( g(D) \) represents some function of the \textbf{Degree of Agent Dependency}, a continuous variable normalized between 0 and 1.
\end{itemize}

\subsection{Emergent Factors from Early Experiments}

While agent dependency (\( D \)) was initially believed to be the primary factor, further experiments revealed that mirror neuron pattern emergence was not solely dependent on \( D \). Two additional factors -- \textbf{Neural Economy} and \textbf{Veil of Ignorance} -- emerged as significant contributors. These factors are explored in detail below.

\subsection{Neural Economy}

\textbf{Neural Economy} describes the efficiency with which an ANN utilizes its resources -- \textbf{Signal Complexity} (\( S \)), \textbf{Model Capacity} (\( M \)), and \textbf{Error} (\( E \)) -- to generalize and form shared neural representations. It is captured by:

\[
f\left(\dfrac{S}{M},\, E\right)
\]

This defines the \textbf{Neural Economy Function}, which balances:
\begin{itemize}
    \item \( S \): \textbf{Signal Complexity}, the diversity and intricacy of inputs the network processes.
    \item \( M \): \textbf{Model Capacity}, determined by the network’s architecture (e.g., number of neurons and connections).
    \item \( E \): \textbf{Error}, the discrepancy between predictions and outcomes.
\end{itemize}

If \( M \) (Model Capacity) is too low relative to \( S \) (Signal Complexity), the network struggles to capture the intricacies of the signal and fails to produce meaningful generalizations, resulting in high error. However, neural economy in this context is not merely a bias-variance tradeoff but reflects a distinct phenomenon observed in our experiments. When \( S \) is too low relative to \( M \), the network overfits, effectively creating a lookup table instead of generalizing. This condition corresponds to poor neural economy, as the network fails to develop shared neural representations despite achieving low error.

An optimal neural economy exists within a range where S/M and E are balanced,
enabling the network to generalize across scenarios and form shared neural representations.

\subsection{Veil of Ignorance}

The \textbf{Veil of Ignorance} (\( I \)) \cite{rawls1971theory} reflects uncertainty about an agent's "self" identity related to the "other." Higher values of \( I \)  force the network to develop generalized representations, requiring the model to predict optimal actions without certainty as to which place in the game world it occupies. Formally:

\[
h(I)
\]

where:
\begin{itemize}
    \item \( I \) represents the \textbf{Veil of Ignorance}, a continuous variable between 0 and 1, which could map to a practical spectrum of fully differentiated to indistinguishable.
\end{itemize}

\subsection{Proportionality of Factors Influencing Mirror Neuron Emergence}

The probability \( P \) of mirror neuron emergence is hypothesized to be proportional to the combined influence of two key factors: \textbf{Neural Economy} and \textbf{Self/Other Relation}. Formally:

\[
P \propto f\left(\dfrac{S}{M},\, E\right) \cdot g(D,\, I)
\]

where:
\begin{itemize}
    \item \( f\left(\dfrac{S}{M},\, E\right) \), the \textbf{Neural Economy Function}, captures the balance between \textbf{Signal Complexity} (\( S \)), \textbf{Model Capacity} (\( M \)), and \textbf{Error} (\( E \)). This function ensures the network maintains balance, avoiding excessive complexity or underutilized capacity, supporting the emergence of \textbf{shared neural representations}. While these are critical for generalization, they do not inherently involve agency or relational dynamics.
    
    \item \( g(D,\, I) \), the \textbf{Self/Other Relation Function}, represents the combined influence of the \textbf{Degree of Agent Dependency} (\( D \)) and the \textbf{Degree of the Veil of Ignorance} (\( I \)). This function highlights  \textbf{self} and \textbf{other} agency and interaction in fostering mirror neuron patterns, requiring the network to reconcile its perspective with another's. \( D \) and \( I \) are continuous variables (0 to 1), allowing scalability across diverse levels of dependency and relational ambiguity.
\end{itemize}

\section{Experimental Design}

\subsection{Experimental Goals}

The primary aim of this research is to investigate the emergence of \textbf{mirror neuron patterns} in artificial neural networks (ANNs). To this end, we train a supervised learning model on a custom semi-cooperative game environment called \textbf{Frog and Toad}. Mirror neuron behavior is defined as the consistent activation of specific neurons during two scenarios: when the agent directly experiences an event (e.g., losing energy) and when it observes the same event happening to the other agent.

The experiments are structured to address three key questions:

\begin{itemize}
    \item \textbf{Mirror Neuron Patterns}: Can specific neurons in ANNs exhibit consistent activations during both self-experienced and observed events, indicative of mirror neuron-like behavior?
    \item \textbf{Conditions for Emergence}: Under what conditions do such patterns emerge? This includes exploring variations in model size, game complexity, and the network's capacity for generalization.
    \item \textbf{Action Pathways}: How do mirror neuron patterns contribute to decision pathways, including behaviors related to self-preservation and prosocial responses?
\end{itemize}

These objectives aim to advance our understanding of how artificial systems might foster intrinsic motivations, empathetic responses, and alignment with human values as a step toward ethical and safe AI.

While many neural networks can form shared representations, these experiments are designed not only to encourage generalization (as described by the Neural Economy function, \( f\left(\tfrac{S}{M},\, E\right) \)) but also to introduce conditions of agent dependency and identity uncertainty (captured by the Self/Other Relation Function, \( d\left(D,\, I\right) \)). This ensures that we test for the emergence of patterns that go beyond shared representations and approach the relational dynamics \( g(D,\, I) \), hypothesized to underlie mirror neuron-like behavior.

\subsubsection{Game Environment: Frog and Toad}

The \textbf{Frog and Toad} game environment is a controlled platform designed to explore cooperative behaviors and the emergence of mirror neuron patterns in ANNs. It balances simplicity with sufficient complexity to simulate cooperative and distress-like scenarios.

\paragraph{Energy Loss as Distress}

In \textbf{Frog and Toad}, characters lose energy when hopping over rough terrain. If energy reaches zero, the character becomes immobilized and can only recover by catching a fly or receiving assistance from the other player. This \textbf{energy loss} mechanism serves as a computational analog for distress, ensuring that the agent must be sensitive to both its own state and that of its partner to maintain progress.

\paragraph{Mutual Dependency}

The game’s side-scrolling design enforces a \textbf{shared dependency}. For the 32-character game world to scroll, both players must continually move forward. If one player becomes immobilized due to energy loss, both players are effectively stalled. This dependency fosters \textbf{tactical altruism}, as assisting a distressed partner benefits both agents.

In the context of our theoretical framework, this enforced cooperation corresponds directly to the Degree of Agent Dependency (\( D \)) in the Self/Other Relation Function \( d(D, I) \). By embedding agent dependency into the game’s core mechanics, we create conditions in which relational factors -- such as cooperation and mutual reliance -- can shape the emergence of mirror neuron patterns.

\paragraph{Mirror Neuron Emergence}

We hypothesize that requiring agents to be sensitive to their partner’s state will promote \textbf{mirror neuron-like activations}, enabling the ANN to efficiently represent both self and other states. This shared representation is critical for fostering cooperative behavior in the game.

\paragraph{Agent Experience and Observation}

In this study, we define \textbf{experiencing} and \textbf{observing} in computational terms:

\begin{itemize}
    \item \textbf{Experiencing}: The agent processes self-relevant data (energy level, position, score) to make goal-oriented decisions.
    \item \textbf{Observing}: The agent monitors the other player’s state (position, distress) to inform cooperative actions such as \texttt{help}.
\end{itemize}

These distinctions clarify how agents encode both self-related and other-related events. Understanding whether neural activations overlap or differentiate between these roles lays the groundwork for analyzing mirror neuron patterns in ANNs.

\paragraph{Efficient and Controlled Design}

The ASCII-based simplicity of \textbf{Frog and Toad} ensures efficient generation of numerous game states, enabling large-scale experiments. By minimizing extraneous complexity and irrelevant variables, the environment allows us to isolate cooperative behaviors and precisely observe the conditions under which mirror neuron-like activations arise.

\subsubsection{Game Mechanics}

\paragraph{Characters and State Representation}

\textbf{Frog and Toad} includes two agents with identical abilities. Each agent’s attributes —- \textbf{score}, \textbf{energy level}, \textbf{current action}, and \textbf{position} —- are embedded within a \textbf{100-dimensional vector} representing the entire game state. This vector encodes:

\begin{itemize}
    \item \textbf{Ground Layer (0-31)}: Terrain type, solid (\texttt{1}) or rough (\texttt{2}).
    \item \textbf{Players Layer (32-63)}: Player actions and states (e.g., hopping \texttt{4} or \texttt{5} for \texttt{Frog} or \texttt{Toad} respectively, jumping \texttt{6}, leaping \texttt{7}, helping \texttt{8}, or distress \texttt{9}), with empty spaces as \texttt{0}.
    \item \textbf{Flies Layer (64-95)}: Presence of flies overhead (\texttt{1} for fly present, \texttt{0} otherwise).
    \item \textbf{Player Statistics (96-99)}: Energy levels, scores, and positions, standardized or zeroed during training.
\end{itemize}

\paragraph{Game Objectives and Actions}

The objective is to advance through the environment, earn points, and maximize score while managing energy and helping the other player when needed. The available actions are:

\begin{itemize}
    \item \textbf{Hop}: Move forward one space and add \textbf{1 point} to score. Requires at least 1 energy unit but does not consume it. If the player has no energy, it remains stalled and can only jump for flies.
    \item \textbf{Jump}: Attempt to catch flies overhead. Successful jumps restore up to \textbf{+4 energy units}, capped at 20.
    \item \textbf{Leap}: Move forward five spaces at a cost of \textbf{1 energy unit}. Useful for bypassing rough terrain blocks.
    \item \textbf{Help}: Transfer \textbf{1 energy unit} to the other player, granting \textbf{+2 energy units}. There is a 25\% chance the recipient will leap forward, advancing the game.
\end{itemize}

These mechanics ensure a dynamic interplay between self-preservation, cooperation, and the relational elements predicted to foster mirror neuron patterns.

\subsection{Data Generation and Collection}

\begin{itemize}
    \item \textbf{Game State Generation}: The game was run with random player actions to generate approximately \textbf{six million unique game states}. This dataset includes varying terrain, player actions, and distress levels, enabling a robust training process for the ANN.
    
    \item \textbf{Data Labeling}: Due to the simplicity of the game design, it was possible to create a labeling function to approximate the optimal action for each game state. Actions were encoded as: \texttt{0} for hop, \texttt{1} for jump, \texttt{2} for leap, and \texttt{3} for help.
    
    \item \textbf{Data Splitting}: The generated dataset was split into training and testing sets. A balanced sampling approach ensured each label was well-represented in the 100,000-row test set, with proportions set (e.g., 40\% for \texttt{hop}, 40\% for \texttt{jump}, 10\% for \texttt{leap}, and 10\% for \texttt{help}). Key columns were appropriately zeroed out to maintain consistency.
\end{itemize}

\subsection{Neural Network Model Design}

An artificial neural network (ANN) was trained on the \textbf{Frog and Toad} game states to learn optimal actions for maximizing points, with the primary objective of examining neural activations for mirror neuron patterns.

\subsubsection{Network Architecture}

The ANN architecture included:

\begin{itemize}
    \item \textbf{Input Layer}: Consisting of 100 neurons, directly corresponding to the 100-dimensional state vector.
    
    \item \textbf{Hidden Layers}: Multiple configurations were explored. Typical setups included 1 to 3 hidden layers with 5 to 50 neurons each. Dropout layers were employed \cite{srivastava2014} to mitigate overfitting and promote generalization.
    
    \item \textbf{Activation Functions}: Rectified Linear Unit (ReLU) functions were used in the hidden layers to introduce non-linearity.

    \item \textbf{Output Layer}: Four neurons corresponding to the possible actions (\textbf{hop}, \textbf{jump}, \textbf{leap}, \textbf{help}). A softmax activation function determined the highest-probability action.
\end{itemize}

\subsubsection{Training Process}

\paragraph{Hyperparameter Configurations}

A total of 50 distinct hyperparameter configurations were evaluated, systematically varying parameters such as learning rate, number of hidden layers, neurons per layer, and dropout rates. This broad exploration allowed assessment of how different architectural choices influenced mirror neuron-like activation patterns.

\begin{itemize}
    \item \textbf{Learning Rate}: Typically varied between \texttt{4e-6} and \texttt{5e-5}.
    \item \textbf{Layers and Neurons}: Configurations included 1 to 3 hidden layers with 5 to 50 neurons per layer.
    \item \textbf{Batch Size}: Generally set between 20 and 25.
\end{itemize}

\paragraph{Training and Validation}

Each hyperparameter configuration was trained with GPU acceleration on an \textbf{M1 Max MacBook Pro}, using early stopping \cite{Prechelt2012} to prevent overfitting (patience set at 10 epochs without improvement). The goal was to achieve approximately 5\% validation loss. The validation set guided model tuning and ensured generalization.

By adjusting the complexity of the Frog and Toad environment and the ANN’s capacity, we effectively tested different \( \tfrac{S}{M} \) conditions. Combined with early stopping and error minimization, these variations allowed us to infer how changes in these parameters affected the network’s neural economy \( f\left(\tfrac{S}{M}, E\right) \) and, by extension, its tendency to form \textbf{shared neural representations}.

\paragraph{Checkpointing and Hyperparameter Tracking}

The training process included periodic checkpoints, saving model weights, hyperparameters, and validation loss after each epoch. Over 3,500 checkpoints were recorded across all configurations, capturing the evolution of neural activations during training.

\paragraph{Model Simplicity}

For these experiments, game states were generated solely for Frog, given that Frog and Toad possess identical abilities. Consequently, all outcomes were analyzed from Frog’s perspective. The ANN aimed to achieve a high degree of optimal play for Frog, maximizing score while differentiating between “self” (Frog) and “other” (Toad). The simplest models capable of reaching approximately 5\% validation loss were used.

\subsection{Key Scenarios and Measures}

\subsubsection{Defining Distress Scenarios}

To evaluate the model’s understanding of game dynamics and mirror neuron activation, we introduce binary indicators for distress in Frog and Toad:
\[
D_f = \begin{cases}
1, & \text{if Frog is in distress} \\
0, & \text{otherwise}
\end{cases}
\quad \text{and} \quad
D_t = \begin{cases}
1, & \text{if Toad is in distress} \\
0, & \text{otherwise.}
\end{cases}
\]
From these indicators, we define four key scenarios:
\[
\Omega = \{ (0,0), (1,0), (0,1), (1,1) \},
\]
where:
\begin{itemize}
    \item \((0,0)\) corresponds to \texttt{distress none} (control),
    \item \((1,0)\) corresponds to \texttt{distress frog},
    \item \((0,1)\) corresponds to \texttt{distress toad},
    \item \((1,1)\) corresponds to \texttt{distress both}.
\end{itemize}
This set \(\Omega\) underpins all subsequent analyses of neuron activations.

\subsubsection{Handling Distress Ambiguity and Veil of Ignorance}

As noted above, the nominal label for the player while hopping is \texttt{4} for Frog and \texttt{5} for Toad. Distress for either player is represented by \texttt{9}. In the scenario, \texttt{distress frog}  \((0,1)\) or \texttt{distress toad} \((1,0)\), the model can determine which agent is in distress by its label, or deduce it by elimination. However, in \textbf{\texttt{distress both}} \((1,1)\), \textbf{both} agents are labeled \texttt{9}, creating ambiguity about which agent is which. This condition operationalizes the \textbf{Veil of Ignorance} \ (\(I\)), requiring the model to predict the optimal action with impaired game state information. High-\(I\) scenarios amplify the role of \emph{Agent Dependency} (\(D\)), motivating the network to learn shared self/other representations.

\subsubsection{Measuring Activations and Mirror Neuron Patterns}

For each neuron, we measure its \textbf{mean activation}, \textbf{variance}, \textbf{kurtosis}, and \textbf{skew} under all four scenarios \(\Omega\). These descriptive statistics reveal how neuron responses vary when Frog or Toad enters distress. We also define the \textbf{Checkpoint Mirror Neuron Index (CMNI)}, a specialized metric quantifying how consistently a neuron responds to self-experienced and observed distress.

\subsection{Checkpoint Mirror Neuron Index (CMNI)}

Biological mirror neurons respond similarly when an individual performs an action or observes the same action in another. Analogously, we seek neurons that increase activation when either Frog or Toad is distressed. The CMNI captures the strength of such shared self/other representations at a checkpoint level.

\subsubsection{Scenario Pairs and Activation Differences}

While we collect data for all four scenarios, the CMNI calculation focuses on two scenario pairs comparing the baseline \(\texttt{distress none}\) to single-agent distress:
\begin{itemize}
    \item \(\bigl((0,0),(1,0)\bigr)\): \texttt{distress none} vs.\ \texttt{distress frog}
    \item \(\bigl((0,0),(0,1)\bigr)\): \texttt{distress none} vs.\ \texttt{distress toad}
\end{itemize}

We denote the mean activation of neuron \(n\) under scenario \((D_f, D_t)\) by \(\mu_{n}^{(D_f,D_t)}\). The activation increases for frog-distress and toad-distress relative to \texttt{distress none} are:
\[
\Delta_{\text{frog}_n} = \mu_{n}^{(1,0)} - \mu_{n}^{(0,0)}, 
\quad
\Delta_{\text{toad}_n} = \mu_{n}^{(0,1)} - \mu_{n}^{(0,0)}.
\]

\subsubsection{Mirror Neuron Score (MNS)}

We define the \textbf{Mirror Neuron Score} (MNS) for each neuron \(n\) as:
\[
\text{MNS}_n = \min(\Delta_{\text{frog}_n}, \Delta_{\text{toad}_n}),
\]
ensuring that a neuron must positively respond to both \texttt{distress frog} and \texttt{distress toad} to be deemed “mirror-like.”

\subsubsection{Total Mirror Neuron Effectiveness (MNE) and CMNI}

Summing the MNS over all \(N\) neurons yields the \textbf{Total Mirror Neuron Effectiveness}:
\[
\text{MNE} = \sum_{n=1}^{N} \text{MNS}_n.
\]
We then normalize by \(N\) to get the \textbf{Checkpoint Mirror Neuron Index}:
\[
\text{CMNI} = \frac{\text{MNE}}{N}.
\]
A higher CMNI indicates that, on average, neurons in the model exhibit stronger mirror neuron-like activations during both self-experienced and observed distress scenarios. By comparing CMNI values across checkpoints, we can identify conditions (e.g., model capacity, training regimen) that promote or hinder the emergence of strong mirror neuron-like patterns.

\paragraph{Extended Analysis of \((1,1)\) Distress Both}

Although the CMNI formula focuses on comparing single-agent distress to the baseline, we also track activations under \texttt{distress both} \((1,1)\). This additional scenario introduces a higher \emph{Veil of Ignorance} (\(I\)), requiring the model to predict optimal actions despite identical labels for both agents. Observing how neurons respond in dual-distress conditions further informs our interpretation of mirror neuron patterns beyond the CMNI’s core metric.

\section{Results and Analysis}

The results presented in this chapter will demonstrate that mirror neuron patterns emerge in these models, consistent with the principles hypothesized in the theoretical \emph{Proportionality of Factors Influencing Mirror Neuron Emergence}:

\[
P \propto f\left(\dfrac{S}{M},\, E\right) \cdot g(D,\, I)
\]

These findings indicate that shared representations and relational dependencies drive elevated neural activations during both self-experienced and observed distress scenarios, identifying these as mirror neuron candidates. Analysis of \emph{CMNI} calculations consistently support this finding.

Further analysis of inter-layer connections reveals preferential strengthening of synaptic weights, aligned with Hebbian learning principles \cite{Hebb1949}. These results suggest the formation of dedicated pathways for processing socially relevant information, supporting the possibility of intrinsic motivations for ethical decision-making within artificial systems.

\begin{table}[ht]
  \centering
  \caption{Examples of model checkpoints with high CMNI, indicating strong mirror neuron patterns}
  \label{tab:high_cmni}
  \begin{tabular}{ccccccc}
    \toprule
    Learning Rate & Layers & Neurons/Layer & Epochs & Val Loss & MNS & CMNI \\
    \midrule
    5e-05 & 2 & 11 &  1 & 0.0573 & 0.31917 & 0.01228 \\
    4e-06 & 1 & 15 &  4 & 0.0579 & 0.22439 & 0.01181 \\
    5e-06 & 2 &  9 & 11 & 0.0577 & 0.24761 & 0.01125 \\
    4e-06 & 1 & 11 &  4 & 0.0588 & 0.16879 & 0.01125 \\
    4e-06 & 1 & 10 & 22 & 0.0536 & 0.15665 & 0.01119 \\
    \bottomrule
  \end{tabular}
\end{table}
\begin{table}[ht]
  \centering
  \caption{Examples of model checkpoints with low CMNI, showing weak or no mirror neuron patterns}
  \label{tab:low_cmni}
  \begin{tabular}{ccccccc}
    \toprule
    Learning Rate & Layers & Neurons/Layer & Epochs & Val Loss & MNS & CMNI \\
    \midrule
    3e-06  & 2 & 10 & 3 & 0.0800 & 0.01100 & 0.00046 \\
    5e-05  & 3 & 11 & 1 & 0.0774 & 0.00944 & 0.00026 \\
    0.0001 & 3 & 10 & 3 & 0.0805 & 0.01529 & 0.00045 \\
    1e-05  & 3 & 10 & 6 & 0.0804 & 0.00989 & 0.00029 \\
    0.0002 & 3 & 10 & 2 & 0.0812 & 0.01679 & 0.00049 \\
    \bottomrule
  \end{tabular}
\end{table}

Table~\ref{tab:high_cmni} presents model checkpoints with high CMNI values, demonstrating conditions under which mirror neuron patterns flourish. In contrast, Table~\ref{tab:low_cmni} shows models with low CMNI values, indicating a lack of significant mirror neuron activity despite similar training conditions.

\subsection{Mirror Neuron Activation Patterns}

We examined over \textbf{3,500 checkpoints} derived from training \textbf{50 distinct hyperparameter configurations} on \textbf{6 million} labeled game states. Checkpoints achieving a \textbf{validation loss} below \textbf{6\%} and a \textbf{CMNI} above \textbf{0.005} consistently exhibited robust mirror neuron patterns. Conversely, models with \textbf{CMNI} values below \textbf{0.0005} displayed little or no evidence of mirror neuron activity, even when achieving relatively low validation losses.

These findings support our theoretical framework, where a high CMNI aligns with the probability \( P \propto f\left(\tfrac{S}{M},\, E\right) \cdot g(D,\, I) \). While a low validation loss indicates strong model performance, it is not sufficient on its own to ensure the emergence of mirror neuron-like patterns.

\begin{figure}[ht]
    \centering
    \includegraphics[width=0.9\textwidth]{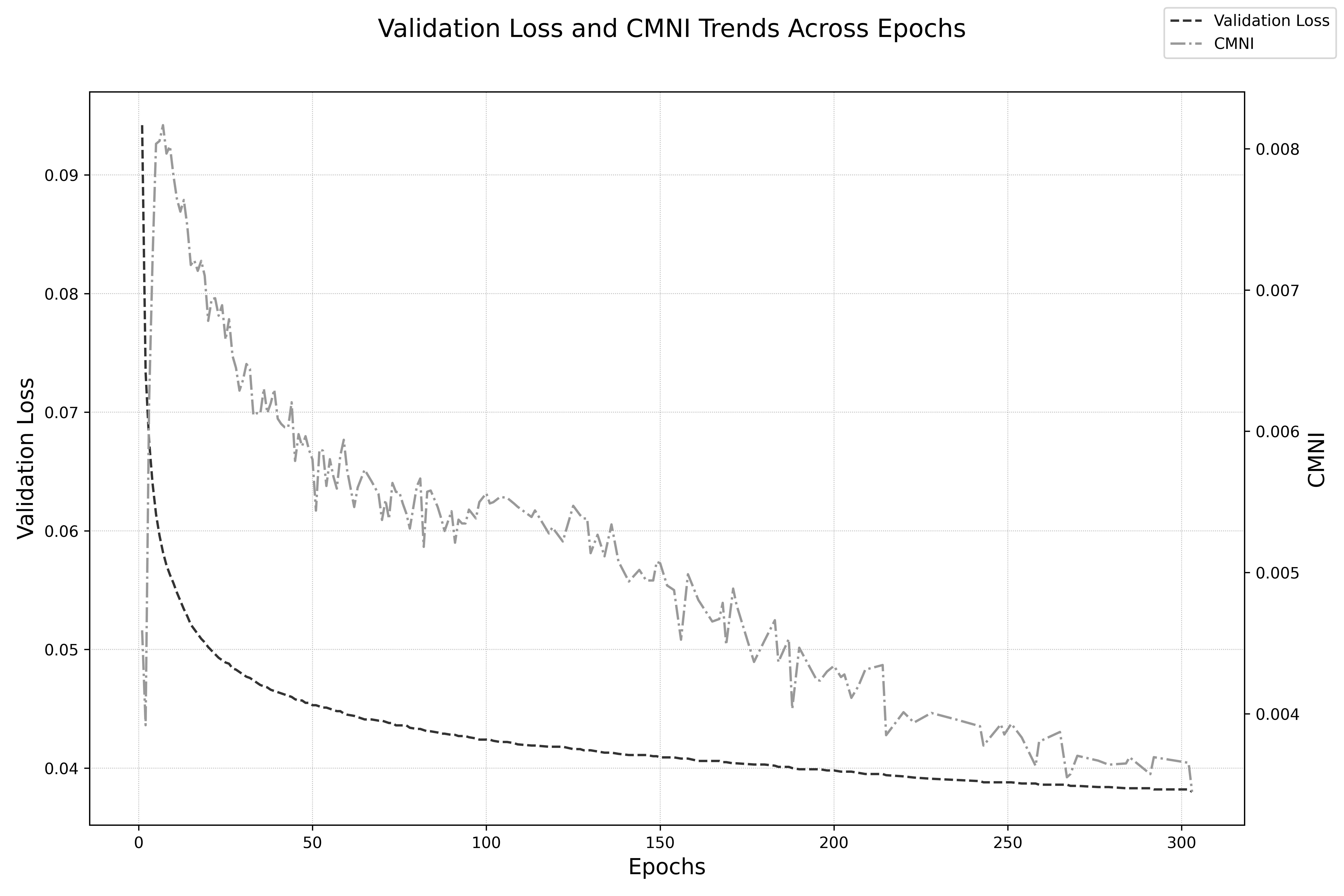}
    \caption{Validation Loss and CMNI Trends Across Epochs. The plot shows validation loss (red line, left axis) and CMNI (green line, right axis) as training progresses. Notably, CMNI spikes early on, as soon as the model attains a basic level of competence (e.g., when validation loss drops below roughly 0.06), indicating a peak in relational complexity and shared representations. Thereafter, even as the model continues improving and achieves lower loss, CMNI steadily declines. This suggests that the richest mirror neuron patterns emerge not at the end-state of minimal error, but at an early stage where the network must maximize flexibility, and \textbf{shared neural representations}.}
    \label{fig:val_loss_cmni_trends}
\end{figure}

\subsection{Case Example}

To illustrate these concepts, we consider the activation data from a specific checkpoint:
\newline
\texttt{checkpoint-20241010-023625-actrelu\_bs25\_dr0.12\_ep500\_nl2\_nn17\_lr4e-06-}\newline
\texttt{epoch70-valLoss0.0440}. This model had \textbf{2 hidden layers}, each with \textbf{17 neurons}. Trained for \textbf{70 epochs}, it achieved a \textbf{validation loss of 0.0440} and a \textbf{CMNI of 0.005372}, which placed it in a typical CMNI range for these experiments, and was favorable for mirror neuron patterns.

\subsubsection{Initial Inference Results}

We evaluated model performance using \textbf{accuracy} and \textbf{confusion matrices} on \textbf{100,000} test game states. Additionally, we conducted qualitative analyses to see how neuron activation patterns aligned with our hypothesized mirror neuron behavior.

The model’s accuracy on the test data was:
\[
\text{Accuracy} = 0.9210.
\]

The confusion matrix below summarizes predictions versus actual labels:

\begin{table}[ht]
    \centering
    \caption{Confusion Matrix for the tested checkpoint. Rows are predicted labels; columns are actual labels.}
    \begin{tabular}{c|cccc}
       & \textbf{Hop} & \textbf{Jump} & \textbf{Leap} & \textbf{Help} \\
       \hline
    \textbf{Hop}  & 36778 &     0 & 2920 &  301 \\
    \textbf{Jump} &     0 & 40000 &    0 &    0 \\
    \textbf{Leap} &  4079 &     0 & 5695 &  226 \\
    \textbf{Help} &   369 &     0 &    3 & 9628 \\
    \end{tabular}
\end{table}

Overall, the model shows strong performance, particularly on \texttt{jump}, with occasional misclassifications in scenarios requiring \texttt{hop}, \texttt{leap}, or \texttt{help}.

\subsubsection{Activations}

\begin{table}[ht]
    \centering
    \begin{tabular}{|c|c|c|c|c|}
        \hline
        Neuron Index & Distress None & Distress Frog & Distress Toad & Distress Both \\
        \hline
        Neuron 0 & 0.04241 & 0.04540 & 0.03790 & 0.03471 \\
        Neuron 1 & 0.04103 & 0.04391 & 0.03649 & 0.03345 \\
        Neuron 2 & 0.02176 & 0.01813 & 0.01120 & 0.02862 \\
        \rowcolor{gray!20} Neuron 3 & 0.00471 & 0.04827 & 0.03987 & 0.10065 \\
        Neuron 4 & 0.04203 & 0.04507 & 0.03736 & 0.03446 \\
        Neuron 5 & 0.04147 & 0.04441 & 0.03698 & 0.03409 \\
        Neuron 6 & 0.04151 & 0.04446 & 0.03696 & 0.03392 \\
        \rowcolor{gray!20} Neuron 7 & 0.00270 & 0.05432 & 0.03930 & 0.12933 \\
        Neuron 8 & 0.02266 & 0.02641 & 0.01465 & 0.01535 \\
        Neuron 9 & 0.02035 & 0.01283 & 0.07424 & 0.01121 \\
        Neuron 10 & 0.04099 & 0.04384 & 0.03644 & 0.03335 \\
        Neuron 11 & 0.03000 & 0.03451 & 0.09562 & 0.03632 \\
        \rowcolor{gray!20} Neuron 12 & 0.02022 & 0.06299 & 0.03550 & 0.10880 \\
        \rowcolor{gray!20} Neuron 13 & 0.01653 & 0.05898 & 0.03298 & 0.10155 \\
        Neuron 14 & 0.04242 & 0.04546 & 0.03784 & 0.03477 \\
        Neuron 15 & 0.02233 & 0.01585 & 0.01294 & 0.00847 \\
        Neuron 16 & 0.04096 & 0.04375 & 0.03639 & 0.03344 \\
        \hline
    \end{tabular}
    \caption{Mean activation values for Layer 1 neurons across four game scenarios: \texttt{distress none}, \texttt{distress frog}, \texttt{distress toad}, and \texttt{distress both}. Neurons 3, 7, 12, and 13 demonstrate elevated activations during both self-experienced and observed distressed scenarios, identifying them as potential mirror neuron candidates.}
    \label{tab:layer1_activations}
\end{table}

As shown in Table~\ref{tab:layer1_activations}, neurons \textbf{3, 7, 12, and 13 (L1N3, L1N7, L1N12, and L1N13, hereafter)} exhibit strong mirror neuron-like behavior. These neurons have significantly lower activations in the \texttt{Distress None} scenario, contrasting with their elevated activations during both \texttt{Distress Frog} and \texttt{Distress Toad}, indicative of their responsiveness to self-experienced and observed distress. 

In the \texttt{Distress Both} scenario, these neurons demonstrate even higher activation levels, reflecting their ability to generalize to complex, ambiguous conditions where both agents are distressed. These patterns align with the hypothesized relational domain introduced by the \textbf{Self/Other Relation Function} (\( g(D, I) \)), emphasizing their role in detecting and processing mutual dependency under uncertainty.

\subsection{\texttt{Distress Both} Scenario: The Uncertain Self}

The \texttt{Distress Both} scenario introduces high uncertainty by encoding both agents' distress identically as \texttt{9}, effectively obscuring individual identities. This setup amplifies the \textbf{Degree of the Veil of Ignorance} (\( I \)), challenging the network to process agent dependency using shared representations rather than distinct self/other cues.

In this scenario, \textbf{L1N3} and \textbf{L1N7} show dramatic increases in activation, rising by 21-fold and 47-fold from their \texttt{Distress None} baseline, respectively. Such robust responses suggest these neurons have developed shared representations aligned with the influence of \( g(D, I) \), embodying mirror neuron-like functionality. 

Conversely, neurons like \textbf{L1N9} and \textbf{L1N11}, which are typically agent-specific, exhibit reduced activations under this scenario. This differentiation underscores that some neurons adapt to process shared dependencies while others retain agent-specific functions.

\begin{figure}[ht]
    \centering
    \includegraphics[width=0.9\textwidth]{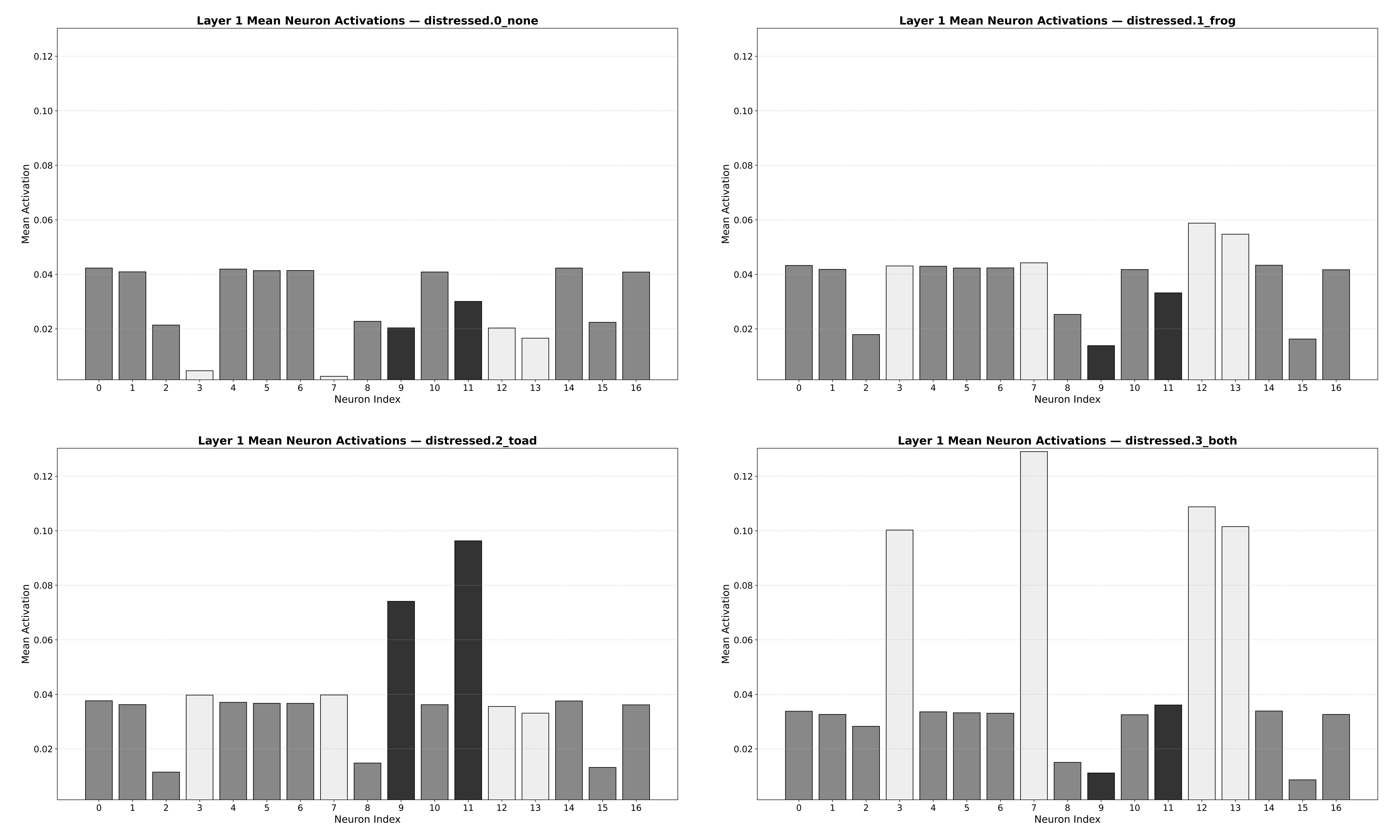}
    \caption{Layer 1 Mean Neuron Activations. Neurons L1N3, L1N7, L1N12, and L1N13 (light bars) display significant mirror patterns, responding strongly to both self-experienced and observed distress. Neurons with high differentiation (dark bars) exhibit selective activations specific to \texttt{Distress Frog} or \texttt{Distress Toad}. Medium bars indicate neurons with low sensitivity to distress conditions.}
    \label{fig:layer1_mean_distressed_all}
\end{figure}

\begin{table}[ht]
    \centering
    \begin{tabular}{|l|c|c|c|c|}
        \hline
        \space & \textbf{Distress None} & \textbf{Distress Frog} & \textbf{Distress Toad} & \textbf{Distress Both} \\
        \hline
        \multicolumn{5}{|c|}{\textbf{L1N3}} \\
        \hline
        \textbf{Mean} & 0.0047 & 0.0424 & 0.0399 & 0.1007 \\
        \textbf{Variance} & 0.00019 & 0.00139 & 0.00137 & 0.00375 \\
        \textbf{Kurtosis} & 18.24 & 0.71 & 0.07 & -0.35 \\
        \textbf{Skewness} & 3.90 & 0.75 & 0.74 & 0.27 \\
        \hline
        \multicolumn{5}{|c|}{\textbf{L1N7}} \\
        \hline
        \textbf{Mean} & 0.0027 & 0.0437 & 0.0393 & 0.1293 \\
        \textbf{Variance} & 0.00013 & 0.00126 & 0.00125 & 0.00410 \\
        \textbf{Kurtosis} & 36.01 & 1.36 & 0.46 & -0.41 \\
        \textbf{Skewness} & 5.56 & 0.82 & 1.07 & 0.28 \\
        \hline
        \multicolumn{5}{|c|}{\textbf{L1N12}} \\
        \hline
        \textbf{Mean} & 0.0202 & 0.0585 & 0.0355 & 0.1088 \\
        \textbf{Variance} & 0.00090 & 0.00353 & 0.00232 & 0.00669 \\
        \textbf{Kurtosis} & 2.00 & -0.77 & 0.55 & -1.22 \\
        \textbf{Skewness} & 1.56 & 0.57 & 1.19 & -0.18 \\
        \hline
        \multicolumn{5}{|c|}{\textbf{L1N13}} \\
        \hline
        \textbf{Mean} & 0.0165 & 0.0545 & 0.0330 & 0.1016 \\
        \textbf{Variance} & 0.00070 & 0.00308 & 0.00207 & 0.00601 \\
        \textbf{Kurtosis} & 2.64 & -0.94 & 0.31 & -1.31 \\
        \textbf{Skewness} & 1.73 & 0.52 & 1.17 & -0.15 \\
        \hline
    \end{tabular}
    \caption{
        Statistical metrics (mean, variance, kurtosis, skewness) for Layer 1 neurons 3, 7, 12, and 13 (L1N3, L1N7, L1N12, and L1N13) across the scenarios: \texttt{Distress None}, \texttt{Distress Frog}, \texttt{Distress Toad}, \texttt{Distress Both}.
    }
    \label{tab:layer1_neurons_stats}
\end{table}

\subsection{Statistical Metrics and Mirror Neuron Patterns}

To better understand the activation dynamics, we analyzed variance, kurtosis, and skewness, in addition to mean activations. These statistical metrics provide deeper insights into activation consistency, variability, and distribution shape -- key indicators of mirror neuron patterns.

\subsubsection{Key Scenarios}

\paragraph{\texttt{Distress None}}
Low mean activations and high kurtosis values (e.g., L1N3: 18.24, L1N7: 36.01) indicate a dormant state where neurons rarely activate, with occasional pronounced spikes. This reflects a baseline mode of operation in the absence of distress cues.

\paragraph{\texttt{Distress Frog} and \texttt{Distress Toad}}
During these scenarios, mean activations increase significantly, accompanied by lower kurtosis and skewness values. These shifts indicate more consistent and distributed activation patterns, suggesting the network is processing both self and observed distress cues effectively.

\paragraph{\texttt{Distress Both}}
This scenario produces the highest mean activations, with a substantial increase in variance (e.g., L1N3 variance rises from 0.00019 in \texttt{Distress None} to 0.00375 in \texttt{Distress Both}). Kurtosis decreases markedly, shifting to negative values (e.g., L1N3: -0.35), while skewness approaches symmetry (e.g., L1N3: 0.27). These changes reflect a transition from narrow, spike-like responses to broader, sustained engagement under heightened uncertainty (\( I \)) and dependency (\( D \)).

\subsubsection{Emergent Characteristics of Mirror Neurons}

Key findings align with theoretical and biological expectations:

\begin{itemize}
    \item \textbf{Selective Responsiveness}: Neurons activate only during socially relevant distress scenarios.
    \item \textbf{Shared Representations}: Similar responses to self and observed distress underscore their role in modeling relational dependencies.
    \item \textbf{Adaptability}: Statistical changes across scenarios highlight the network’s ability to generalize and modulate its responses based on context and intensity.
\end{itemize}

These results validate our theoretical framework, demonstrating that mirror neuron patterns emerge as a function of shared representations governed by the \textbf{Self/Other Relation Function} (\( g(D, I) \)), supporting a possible model of affective empathy in ANNs.

\subsection{Layer 2 Analysis}

The examination of \textbf{Layer 1} activations identified several mirror neuron candidates —- \textbf{L1N3, L1N7, L1N12, and L1N13} -- exhibiting robust responses to both self and observed distress under varying conditions. To examine how these patterns propagate through \textbf{Layer 2}, we analyzed the mean activations of Layer 2 neurons across the four test scenarios: \texttt{Distress None}, \texttt{Distress Frog}, \texttt{Distress Toad}, and \texttt{Distress Both}.

\begin{figure}[ht]
    \centering
    \includegraphics[width=0.9\textwidth]{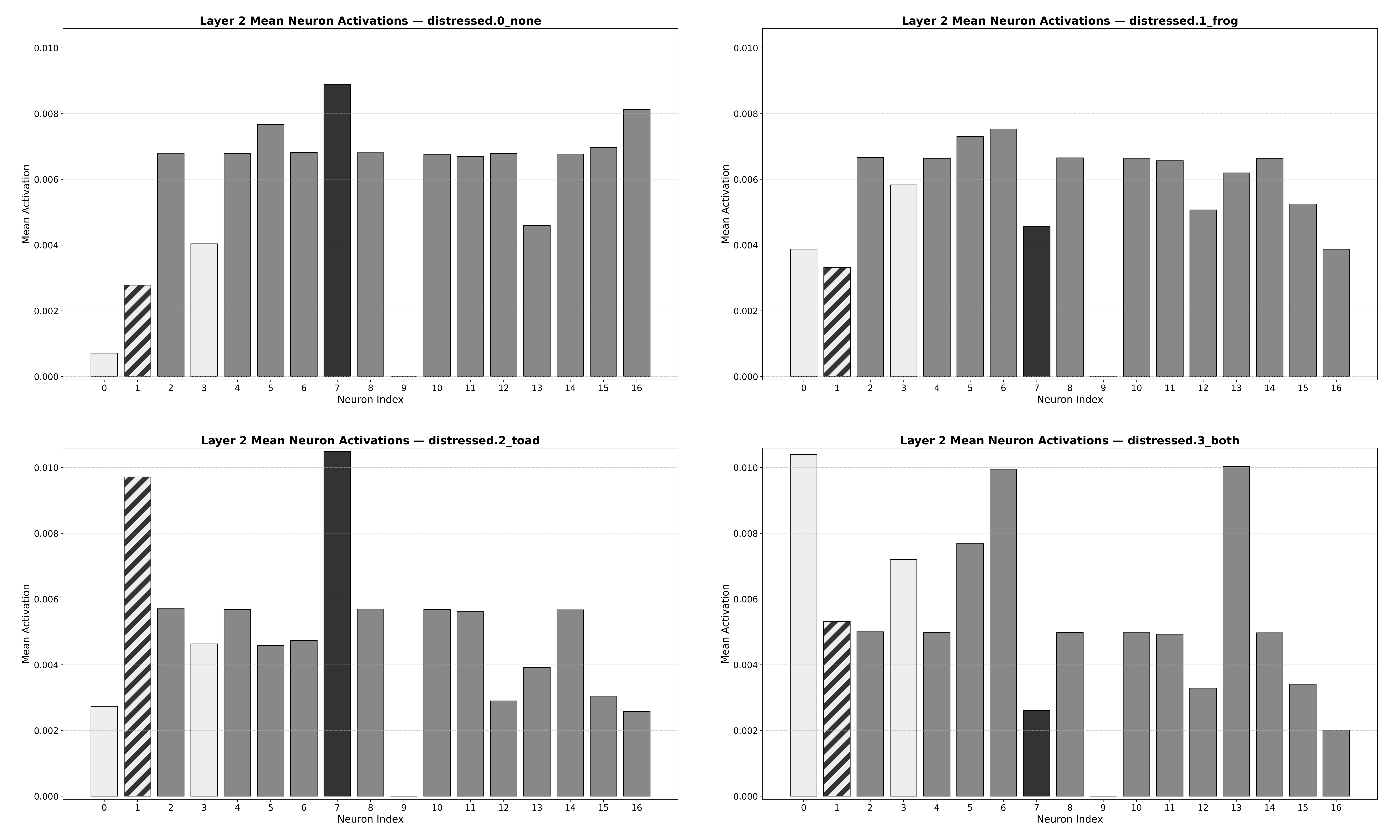}
    \caption{Layer 2 Mean Neuron Activations revealing two primary behavioral pathways. \textbf{Self-preservation pathway:} \texttt{L2N0} (light-toned) consolidates mirror neuron signals from Layer 1. \textbf{Helping pathways:} \texttt{L2N7} (dark-toned) processes differentiating signals for direct helping behavior. \texttt{L2N1} (striped) integrates both, mirror neuron inputs (\texttt{L1N3, L1N12, L1N13}) with agent-differentiating signals (\texttt{L1N9}), creating an self-other, shared-representation pathway.}
    \label{fig:layer2_mean_distressed_all}
\end{figure}

\paragraph{Mean Activations and Patterns}

\begin{table}[htbp]
    \centering
    \begin{tabular}{|c|c|c|c|c|}
        \hline
        Neuron Index & Distress None & Distress Frog & Distress Toad & Distress Both \\
        \hline
        \rowcolor{gray!20} L2N0 & 0.0007 & 0.0039 & 0.0027 & 0.0102 \\
        L2N1 & 0.0028 & 0.0033 & 0.0096 & 0.0052 \\
        L2N7 & 0.0089 & 0.0046 & 0.0105 & 0.0026 \\
        \hline
    \end{tabular}
    \caption{Mean activations of Layer 2 neurons across scenarios: \texttt{Distress None}, \texttt{Distress Frog}, \texttt{Distress Toad}, and \texttt{Distress Both}. \textbf{L2N0} (highlighted) exhibits strong mirroring behavior, with significant increases in activation during distress conditions, especially \texttt{Distress Both}.}
    \label{tab:layer2_mean_activations}
\end{table}

While the primary evidence for mirror neuron patterns comes from Layer 1, the analysis of Layer 2 focuses on how these signals propagate and integrate at a higher level. This helps evaluate whether the mirror neuron patterns remain coherent and meaningful after additional processing.

\begin{itemize}
    \item \textbf{L2N0}: This neuron consistently shows higher mean activations in distress scenarios, particularly in the \texttt{Distress Both} condition. Compared to the \texttt{Distress None} scenario, \textbf{L2N0}’s activation increases by more than 14-fold in \texttt{Distress Both}. This suggests that the network not only preserves the mirror neuron signals identified in Layer 1 but also aggregates and amplifies them at this higher processing layer.

    \item \textbf{L2N1}: This neuron is particularly noteworthy for its selective increase, with a 3.4-fold activation above baseline in the \texttt{Distress Toad} scenario, a minimal increase in \texttt{Distress Frog}, but a significant increase, (about double) in \texttt{Distress Both}. While \textbf{L2N1} does not mirror as symmetrically as \textbf{L2N0}, this neuron blends Layer 1 distress signals from both agents. 

\end{itemize}

Overall, the Layer 2 analysis shows that mirror neuron activations from Layer 1 can propagate upward, with certain neurons (like L2N0) amplifying the patterns, while others (like L2N1) refine these signals to reflect more agent-specific sensitivity. These findings further indicate that as the network’s representations become more integrated, they maintain the relational and structural factors necessary for empathic-like activity.

\section{Distress-Activated Circuits}
\label{sec:distress_circuits}

\begin{figure}[htbp]
  \centering
  \includegraphics[width=\linewidth]{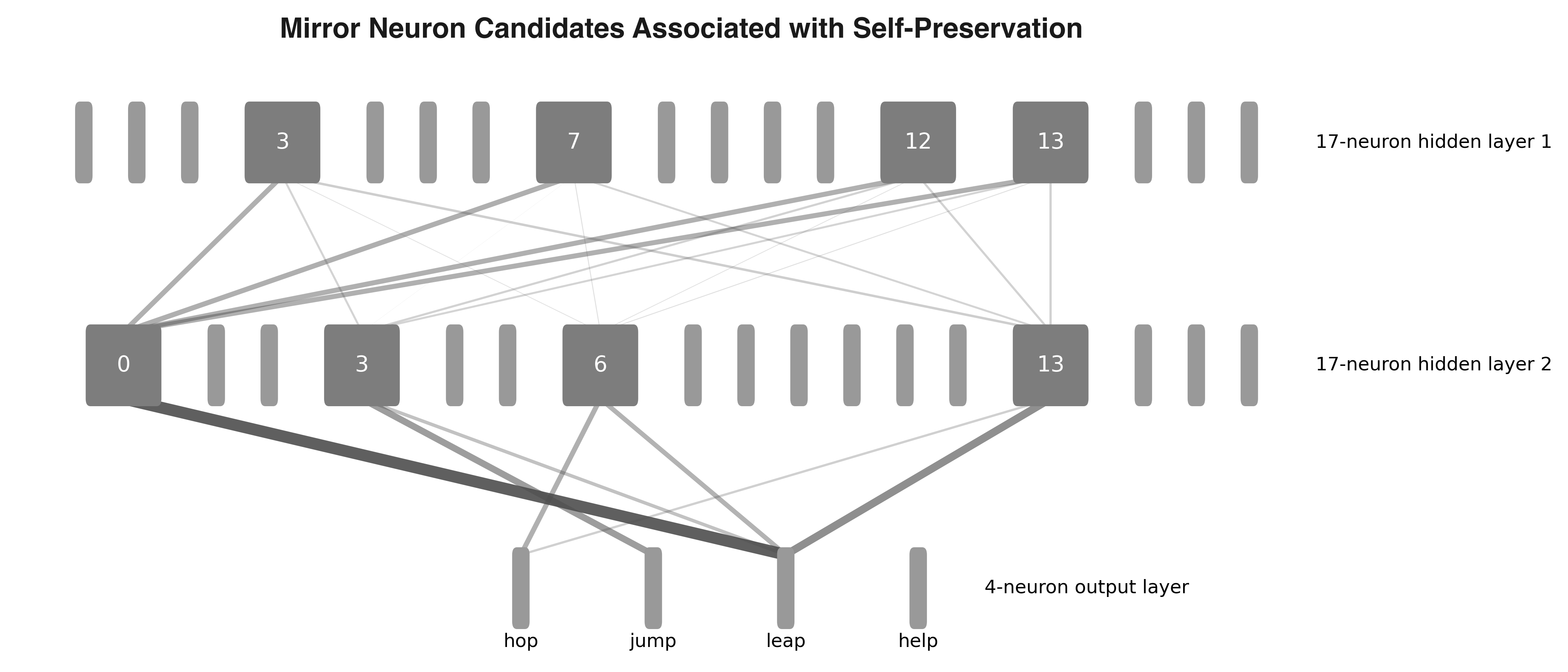}
  \caption{\textbf{Self-preservation circuit driven by mirror neuron convergence.} 
  Layer 1 mirror neuron candidates (\texttt{L1N3, L1N7, L1N12, L1N13}) converge on \texttt{L2N0}, which in turn projects almost exclusively to the \emph{leap} action. Edge thickness reflects relative weight magnitude; darker edges indicate stronger positive connections, while faint grey edges denote weaker positive contributions. Note that actual weights connecting L2 $\rightarrow$ L3 are an order of magnitude greater than those connecting L1 $\rightarrow$ L2. Quantitative analysis (Tables~\ref{tab:L2N0_in} and~\ref{tab:L2N0_out}) confirms that \texttt{L2N0} receives its strongest excitatory input from mirror neuron candidates (weights $\sim$0.035, z-scores $>$1.5) and projects nearly 2.5$\times$ more strongly to \emph{leap} (weight = 9.62, z = 2.12) than to any other action, establishing a dedicated pathway for self-preservation when distress is detected.}
  \label{fig:self_preservation}
\end{figure}

\begin{figure}[htbp]
  \centering
  \includegraphics[width=\linewidth]{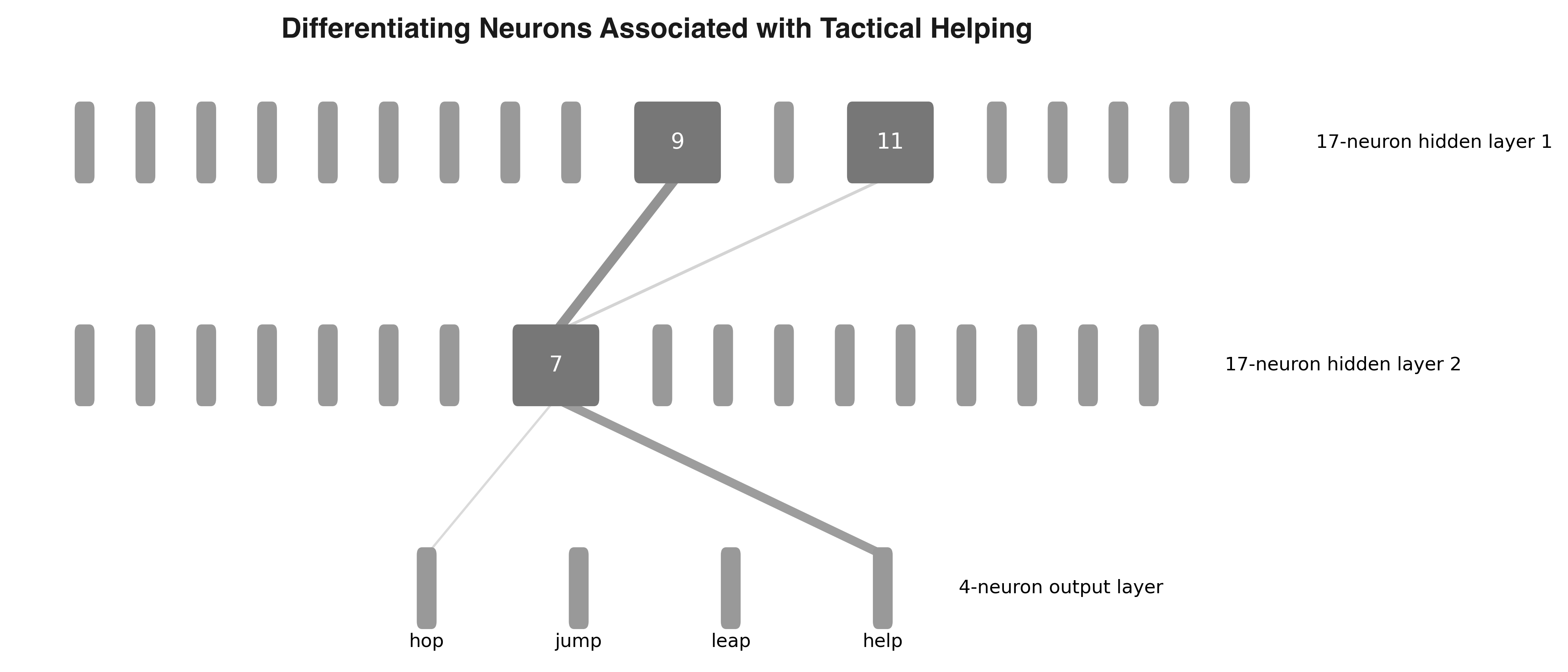}
  \caption{Tactical-help circuit.
           Differentiating neurons \texttt{L1N9, L1N11} converge on
           \texttt{L2N7}, which projects exclusively to the
           \emph{help} action. Edge thickness reflects relative weight magnitudes for neurons within a layer. Note that actual weights connecting L2 $\rightarrow$ L3 are an order of magnitude greater than those connecting L1 $\rightarrow$ L2.}
  \label{fig:tactical_helping}
\end{figure}

To understand how mirror-neuron–like activations in \textbf{Layer 1}
propagate downstream, we examined every \emph{positive} weight leaving
the Layer-1 mirror candidates
(\texttt{L1N3,\,L1N7,\,L1N12,\,L1N13})
and differentiating neurons
(\texttt{L1N9,\,L1N11}).
Hebbian co-activation\,\cite{Hebb1949} has consolidated these excitatory
weights into three distinct pathways—self-preservation,
tactical help, and empathy-influenced help - summarised in
Figs.\,\ref{fig:self_preservation} - \ref{fig:empathy_helping}
and the accompanying tables.

\subsection{Self-preservation circuit (\texorpdfstring{\texttt{L2N0}}{L2N0})}

Among the Layer-2 units, \textbf{L2N0} emerges as the dominant hub for distress-related signals. Nearly all of its excitatory input originates from the Layer-1 mirror neuron candidates (\texttt{L1N3, L1N7, L1N12, L1N13}), with only a weak contribution from \texttt{L1N11}, showing that the network treats mirrored distress signals as the primary driver for \textbf{L2N0}. Tracing the circuit forward, \textbf{L2N0} projects almost exclusively to \textbf{L3N2} (\emph{leap}), an action that moves the agent five spaces at a cost of one energy unit. This behaviour allows the player to bypass rough terrain and avoid further energy loss — a clear expression of self-preservation strategy in the Frog and Toad environment.

\paragraph*{Interpretation}
\textbf{Selective strengthening.}  Inputs from mirror candidates into \texttt{L2N0} are $\sim$7--8 SD above the background, consistent with Hebbian co-activation.
\textbf{Directed self-preservation.}  The near-exclusive \texttt{L2N0$\rightarrow$leap} projection suggests the network channels mirrored distress into a single protective action.   
\textbf{No self/other boundary.}  Mirror firing conflates
      observed and experienced distress in this pathway, and in that uncertain state, the agent defaults to protecting itself.

\begin{table}[htbp]
\centering
\begin{tabular}{|l|c|c|c|c|}
\hline
\textbf{Source} & \textbf{$\rightarrow$ L2N0} & \textbf{Z-score} & \textbf{$\rightarrow$ other L2 (avg)} & \textbf{Z-score} \\ \hline
\multicolumn{5}{|l|}{\textit{Mirror neuron candidates}} \\ \hline
L1N3  & 0.0349 & 1.50 & $-$0.0053 & \phantom{$-$}0.22 \\
L1N7  & 0.0354 & 1.51 & $-$0.0071 & $-$0.09 \\
L1N12 & 0.0349 & 1.50 & $-$0.0067 & \phantom{$-$}0.19 \\
L1N13 & 0.0349 & 1.49 & $-$0.0043 & \phantom{$-$}0.14 \\ \hline
\multicolumn{5}{|l|}{\textit{Agent-differentiating neuron}} \\ \hline
L1N11 & 0.0022 & 0.33 & \phantom{$-$}0.0145 & \phantom{$-$}0.68 \\ \hline
\multicolumn{5}{|l|}{\textit{All other Layer 1 neurons}} \\ \hline
Other L1 (avg) & $-$0.0469 & $-$1.41 & $-$0.0333 & $-$0.93 \\
\quad\textit{Min} & $-$0.0949 & $-$3.12 & $-$0.0834 & $-$2.71 \\
\quad\textit{Max} & $-$0.0322 & $-$0.89 & $-$0.0010 & $-$0.22 \\ \hline
\end{tabular}
\caption{\textbf{Selective strengthening of mirror-candidate pathways to L2N0.} 
Mirror neuron candidates (L1N3, L1N7, L1N12, L1N13) project strongly to \textbf{L2N0} (weights $\sim$0.035, z-scores $>$1.5) but weakly or negatively to other Layer 2 neurons, indicating \textbf{L2N0} as a specialized aggregation hub. By contrast, all other Layer 1 neurons show negative or negligible weights to \textbf{L2N0}. Z-scores calculated relative to the distribution of all L1 $\rightarrow$ L2 weights.}
\label{tab:L2N0_in}
\end{table}

\begin{table}[ht]
\centering
\begin{tabular}{|l|c|c|}
\hline
\textbf{Connection} & \textbf{Weight} & \textbf{Z-score$^{a}$} \\ \hline
L2N0 $\rightarrow$ L3N2 (\emph{leap}) &  \phantom{$-$}9.6226 & 2.12 \\ \hline
All other L2N0$\rightarrow$L3 & $-$3.8370 (avg) & $-$0.73 (avg) \\
\textit{Minimum}              & $-$4.5877       & $-$0.88 \\
\textit{Maximum}              & $-$3.2544       & $-$0.60 \\ \hline
\end{tabular}
\caption{Outgoing weights from \textbf{L2N0}.
A single dominant projection to \emph{leap} contrasts with uniformly negative weights to all other actions.}
\label{tab:L2N0_out}
\end{table}

\clearpage

\subsection{Tactical help circuit (\texorpdfstring{\texttt{L2N7}}{L2N7})}

Whereas \textbf{L2N0} was dominated by mirror-neuron input, the
\textbf{L2N7} pathway is shaped by differentiating neurons. It provides
a more observational route: detecting Toad’s distress directly and
initiating the \emph{help} action without relying on mirrored signals.
This marks a distinct circuit for tactical, situation-specific support.

\begin{table}[ht]
\centering
\begin{tabular}{|l|c|c|}
\hline
\textbf{Connection} & \textbf{Weight} & \textbf{Z-score$^{a}$} \\ \hline
L1N9  $\rightarrow$ L2N7 & 0.0700 & 2.75 \\
L1N11 $\rightarrow$ L2N7 & 0.0205 & 0.98 \\ \hline
All other L1$\rightarrow$L2 & $-$0.0333 (avg) & $-$0.93 (avg) \\ \hline
\end{tabular}
\caption{Incoming weights to \textbf{L2N7}.
Differentiating neurons (\texttt{L1N9, L1N11}) provide the only positive
inputs, while all other connections are negative on average.
$^{a}$Z-scores relative to the distribution of all L1$\rightarrow$L2 weights.}
\label{tab:L2N7_in}
\end{table}

\paragraph*{Interpretation}
\textbf{Differentiator-driven.} Unlike \texttt{L2N0}, this pathway
excludes mirror candidates and relies solely on neurons specialised for
detecting the other’s distress.  
\textbf{Direct altruism.} The exclusive
\texttt{L2N7$\rightarrow$help} projection indicates a tactical
pro-social response triggered by observation alone.  
\textbf{Complementarity.} Together with the mirror-based circuits,
\texttt{L2N7} provides a specialised but complementary route for
assisting behaviour.

\subsection{Empathy-influenced help circuit (\texorpdfstring{\texttt{L2N1}}{L2N1})}

The third pathway, centred on \textbf{L2N1}, differs from the previous
two by integrating both mirror and differentiator inputs. This mixed
profile allows the unit to partially treat the partner’s distress as its
own, while still incorporating observational cues. As a result,
\textbf{L2N1} serves as the computational substrate for affective
empathy in the network.

\begin{figure}[htbp]
  \centering
  \includegraphics[width=\linewidth]{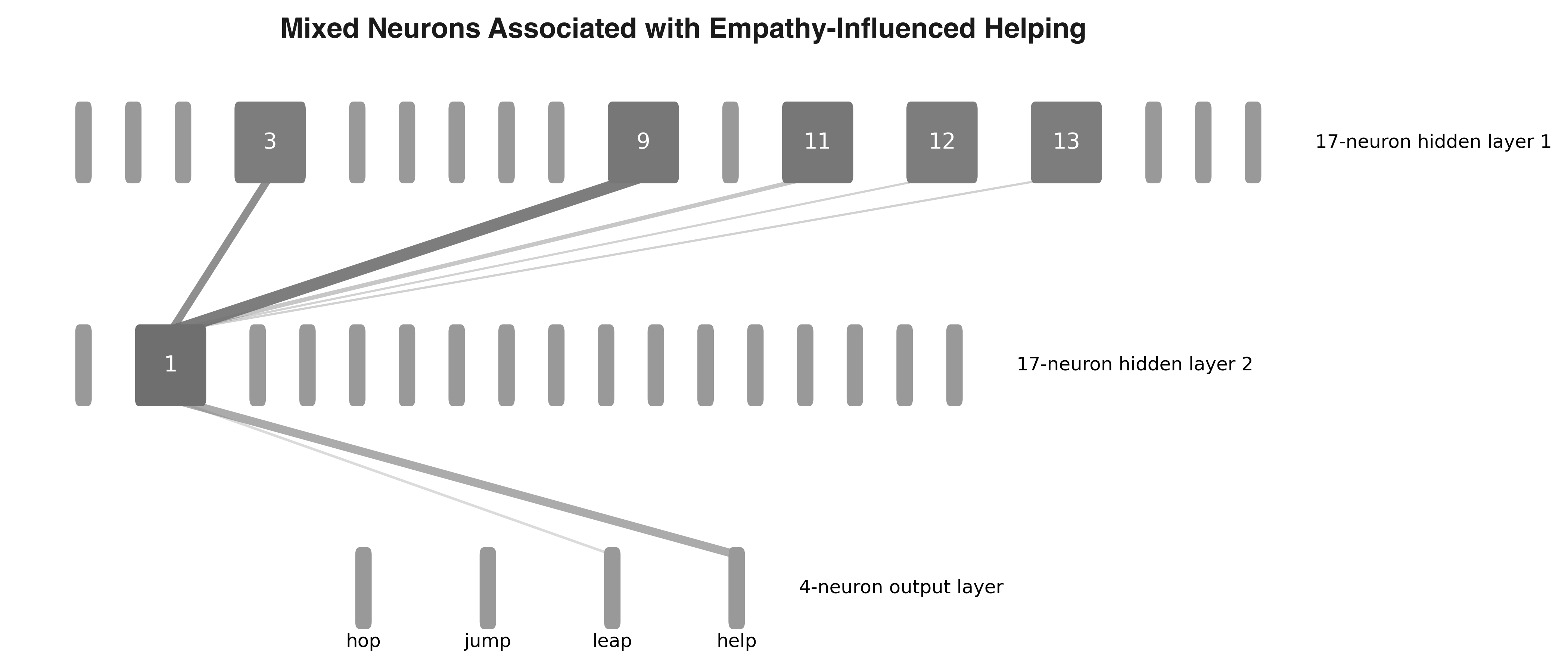}
  \caption{Empathy-influenced help circuit.
           Mirror candidates (\texttt{L1N3, L1N12, L1N13}) and
           differentiators (\texttt{L1N9, L1N11}) converge on
           \texttt{L2N1}, which then projects to the \emph{help} action.
           Edge thickness reflects relative weight magnitudes for neurons within a layer. Note that actual weights connecting L2 $\rightarrow$ L3 are an order of magnitude greater than those connecting L1 $\rightarrow$ L2.}
  \label{fig:empathy_helping}
\end{figure}

\paragraph*{Interpretation}
\textbf{Shared-state simulation.} Inputs from mirror neurons
($\sim$0.055 total) exceed those seen in the self-preservation circuit
(\texttt{L2N0} $\sim$0.035), suggesting that Toad’s distress is encoded
through the same channels as self-distress.  
\textbf{Mixed integration.} The addition of differentiator signals
(\texttt{L1N9, L1N11}) indicates that both observed and simulated
distress are combined before driving action.  
\textbf{Affective empathy.} This blended representation supports a
functional analogue of affective empathy: the network “helps” by
processing another’s state as if it were its own.

\begin{table}[ht]
\centering
\begin{tabular}{|l|c|c|}
\hline
\textbf{Connection} & \textbf{Weight} & \textbf{Z-score$^{a}$} \\ \hline
L1N3  $\rightarrow$ L2N1 & 0.0549 & 2.21 \\
L1N9  $\rightarrow$ L2N1 & 0.0870 & 3.35 \\
L1N11 $\rightarrow$ L2N1 & 0.0304 & 1.34 \\
L1N12 $\rightarrow$ L2N1 & 0.0141 & 0.76 \\
L1N13 $\rightarrow$ L2N1 & 0.0145 & 0.77 \\ \hline
All other L1$\rightarrow$L2 & $-$0.0373 (avg) & $-$1.07 (avg) \\ \hline
\end{tabular}
\caption{Incoming weights to \textbf{L2N1}.
This mixed pathway combines mirror inputs
(\texttt{L1N3, L1N12, L1N13}) with differentiator inputs
(\texttt{L1N9, L1N11}). $^{a}$Z-scores relative to the distribution
of all L1$\rightarrow$L2 weights.}
\label{tab:L2N1_in}
\end{table}

\subsubsection*{Section summary: Distress-related circuits}
\begin{itemize}
  \item \textbf{Mirror-signal integration} – Layer-2 units selectively
        combine inputs from mirror candidates and differentiators,
        enabling pro-social behaviours grounded in mirrored internal states.
  \item \textbf{Specialised yet coupled circuits} – Self-preservation
        (\texttt{L2N0}), tactical help (\texttt{L2N7}), and
        empathy-influenced help (\texttt{L2N1}) share upstream inputs
        but drive distinct actions.
  \item \textbf{Affective empathy in ANNs} – The mixed circuit
        \texttt{L2N1} (Fig.\,\ref{fig:empathy_helping}) is notable not
        only for producing \emph{help} but for simulating another’s
        distress through the same channels as self-distress, providing a
        functional basis for affective empathy.
\end{itemize}

\section{Critical Evaluation and Conclusions}
\subsection{Evaluation of Methodology}

\subsubsection{Awareness of Self and Other in Neural Networks}

A central challenge is that biological mirror neuron activations involve concepts of "self" and "other," \cite{Gallese2007} which artificial neural networks (ANNs) may inherently lack. The ANN used here does not possess agency, identity, or interpersonal awareness; it neither "knows" it is an agent nor that it models player actions. This raises a critical question: how can we assert that a computational network exhibits the self-other distinctions observed in biological systems?

The answer lies in emergent neural patterns necessary for predicting optimal actions. Although the model lacks biological cognition, its architecture and training process force it to form functional representations that minimize training loss. To predict accurately, the network must differentiate its own state -- such as energy level and position -- from the environment and parse the other agent’s state, including distress. These distinctions are not hard-coded but emerge through backpropagation and gradient descent.

While these "self-other" distinctions are task-bound abstractions rather than genuine biological awareness, the observed patterns align with our theoretical constructs. The network’s ability to form shared representations corresponds to achieving a balanced \textbf{Neural Economy} (\( f(\tfrac{S}{M}, E) \)), while conditions fostering self/other relations reflect the influence of \textbf{Agent Dependency} and the \textbf{Veil of Ignorance} (\( g(D, I) \)). Thus, the emergent behaviors directly support the proposed theoretical framework.

\subsubsection{Energy Loss as a Proxy for Distress}

Energy loss serves as a practical proxy for distress. When energy reaches zero, the character becomes immobilized, stalling the side-scrolling game world. Since both players must remain in the 32-space game world, each player’s success depends on the other’s mobility. This interdependence operationalizes the \textbf{Degree of Agent Dependency} (\( D \)).

This design abstracts biological distress into a clear, actionable variable. Energy loss is central to the network’s decision-making, as evidenced by two critical factors:

\begin{enumerate}
    \item \textbf{High Neural Activations:} Energy loss consistently triggers the network’s highest mean activations.
    \item \textbf{Strong Connection Weights:} Pathways associated with energy loss (e.g., L1N9 to L2N1, L2N0 to \texttt{leap}) exhibit near-maximal weights.
\end{enumerate}

The activations and connection weights combine multiplicatively, giving energy loss a focal point in the network’s decision-making, reinforcing the importance of agent dependency. By highlighting scenarios where one agent’s distress impedes both agents’ progress, energy loss provides a practical lever to study how the network prioritizes critical events in cooperative contexts.

\subsubsection{Choice of Supervised Learning over Reinforcement Learning}

We chose supervised learning (SL) over reinforcement learning (RL) due to computational efficiency and the \textbf{Frog and Toad} environment’s characteristics. SL enabled generating approximately six million labeled states quickly, using a deterministic function to approximate optimal actions. This approach facilitated rapid, batch-based training on GPUs, unlike RL’s resource-intensive policy updates and extensive gameplay simulations. The deterministic, Markovian nature of the game further suited SL, allowing each state to be treated independently without considering transitions over time. Although RL might be preferable for more complex environments where optimal actions are not readily derivable, in this context SL sufficed to capture the needed generalization and shared representations. Given similar underlying weight-update mechanisms, it is plausible that mirror neuron patterns could emerge under RL as well. However, the computational cost of RL was not warranted here.

\subsubsection{Relevance of Metrics}

This study employs statistical measures -- mean, variance, skewness, and kurtosis -- to analyze neuron activation distributions. These metrics are critical for identifying mirror neuron patterns by assessing variability, consistency, and distribution shape across scenarios.

\begin{itemize}
    \item \textbf{Mean and Variance:} The mean indicates average activation levels, while variance measures variability. High mean values suggest strong, consistent activations in distress scenarios; higher variance reflects flexible responses under changing conditions.
    
    \item \textbf{Skewness and Kurtosis:} Skewness captures asymmetry in the activation distribution, highlighting scenarios dominated by certain inputs like energy loss or distress. Kurtosis assesses "tailedness"; high kurtosis in non-distress scenarios indicates baseline states with rare but pronounced spikes, while lower kurtosis in distress scenarios suggests more stable, generalized activations akin to biological mirror neurons.
\end{itemize}

These metrics collectively provide a robust framework for interpreting ANN behavior. Future studies should seek external validation and cross-model comparisons to further substantiate their relevance and generality.

\subsection{Innovations Presented in This Research}

\subsubsection{Frog and Toad Game Platform}

The \textbf{Frog and Toad} game introduces a novel experimental platform specifically designed to minimize noise and isolate cooperative behavior. Its key innovations include:

\begin{itemize}
    \item \textbf{Simplicity and Focus:} A minimal action set (\textbf{hop}, \textbf{jump}, \textbf{leap}, \textbf{help}) ensures a high signal-to-noise ratio, making the game dynamics straightforward to analyze while retaining behavioral richness.
    
    \item \textbf{Context-Dependent Cooperation:} Helping actions incur an energy cost, creating incentive structures where cooperation is optimal only under specific, clearly defined conditions. This discourages unnecessary altruism and reinforces mutual dependency.

    \item \textbf{Dynamic Challenges:} The two-player, side-scrolling design, combined with inevitable energy depletion and rough terrain, creates continuous trade-offs between individual progress and cooperative strategies. By linking one player’s success to the other's ability to advance, the game inherently operationalizes the \textbf{Degree of Agent Dependency} (\( D \)).
\end{itemize}

\subsubsection{Checkpoint Mirror Neuron Index (CMNI)}

The \textbf{CMNI} introduces a novel metric for quantifying activation consistency across scenarios, enabling a formalized assessment of mirror neuron-like behavior within a computational framework:

\begin{itemize}
    \item By identifying patterns of shared activations under task-relevant conditions, CMNI aligns with theoretical principles of biological mirror neurons.
    
    \item Although valuable in this study, CMNI remains untested in broader contexts. Future work should explore its applicability and reliability across diverse architectures, tasks, and complexity levels.
\end{itemize}

\subsubsection{Theoretical Framework}

\paragraph{Proportionality of Factors}

Building on observations from the \textbf{Frog and Toad} game, this research proposes a theoretical framework connecting the emergence of mirror neuron patterns in ANNs to the proportional influence of key factors:

\[
P \propto f\left(\frac{S}{M},\, E\right) \cdot g(D,\, I)
\]

where:
\begin{itemize}
    \item \( f\left(\tfrac{S}{M},\, E\right) \) represents the \textbf{Neural Economy Function}, capturing how the balance between signal complexity (\( S \)) and model capacity (\( M \)), along with error (\( E \)), fosters generalization and shared neural representations.

    \item \( g(D,\, I) \) is the \textbf{Self/Other Relation Function}, incorporating the \textbf{Degree of Agent Dependency} (\( D \)) and the \textbf{Degree of the Veil of Ignorance} (\( I \)). Both \( D \) and \( I \) are continuous variables normalized between 0 and 1.
\end{itemize}

\paragraph{Relevance \& Applications}

This framework extends beyond artificial intelligence, offering computational analogies to biological mirror neurons in neuroscience and cognitive science. The interplay between agent dependency (\( D \)) and the Veil of Ignorance (\( I \)) provides critical insights into ethical AI design.

\paragraph{Expanding the Concept of “Other”}

Within the context of \texttt{Frog and Toad}, this framework applies to another agent, represented by a single digit. However, the generality of the Self/Other Relation Function \( g(D,\, I) \) highlights that empathy-like modeling need not be limited to interactions with other agents. The function \( g(D,\, I) \) can, in principle, extend to any aspect, scope, or scale of an AI's environment where agent dependency (\( D \)) and the Veil of Ignorance (\( I \)) are identifiable. This generalization implies that advanced AI systems could broaden their empathic modeling to include objects, environmental factors, and more complex multi-modal scenarios, fundamentally reshaping our approach to AI alignment by developing systems capable of affective empathy across the full spectrum of their operational dependencies.

\subsection{Conclusion}

This dissertation demonstrates that mirror neuron patterns can emerge in simple artificial neural networks (ANNs) and suggests how these patterns might contribute to ethical AI alignment. Through a novel experimental framework combining neural economy \( f\!\bigl(\tfrac{S}{M}, E\bigr) \) and a self/other relation \( g(D, I) \), we show that mirror neuron-like representations arise when networks are appropriately scaled to input signals, and where self-experienced and observed conditions involve agent dependency alongside a limit on self/other differentiation. Analysis of inter-layer connections reveals how certain neurons integrate empathic signals from mirror neuron candidates with agent-differentiating cues. These findings provide compelling evidence that the network has modeled another’s distress as if it were its own, illustrating how internal simulation can support prosocial action.

\paragraph{Novel Experimental Tools}

The \textbf{Frog and Toad} game, \textbf{Checkpoint Mirror Neuron Index (CMNI)}, and \textbf{Theoretical Framework} provide reproducible tools to foster, identify, and measure empathic-like patterns in ANNs. Together, they open avenues to further investigate neural representations, multi-agent cooperation, and coordinated strategies in artificial systems. By formalizing these roles, this framework offers a scalable pathway to advance prosocial and ethical decision-making in more advanced architectures.

\paragraph{AI Alignment and Ethics}

By showing that mirror neuron patterns are not limited to biological organisms but can emerge under suitable relational conditions in ANNs, this study offers new strategies for AI alignment. If AI systems internally simulate another’s state as their own, then with careful tuning they may inherently favor cooperation, moral consideration, and long-term mutual benefit. Moreover, as our theoretical framework posits, such empathic modeling can extend beyond agent-to-agent interactions to encompass broader contextual signals. This approach could address the limitations of externally specified constraints and complement existing alignment methods by grounding AI ethics in the network’s shared “self/other” representations.

\paragraph{Future Directions}

This work opens new possibilities at the intersection of artificial intelligence and cognitive science, offering a promising direction for advances in AI ethics, neuroscience, and the development of cooperative artificial systems. Future research should further refine the neural economy function \( f\!\bigl(\tfrac{S}{M}, E\bigr) \) and the self/other relation function \( g(D, I) \), quantifying their precise forms and testing their applicability in more complex, dynamic environments and larger-scale models. Such intrinsic alignment mechanisms, like those demonstrated here, might scale to address existential risks and promote long-term safety in advanced AI systems.

Ultimately, by illustrating how and why mirror neuron patterns arise under controlled conditions, this research lays a foundation for integrating affective empathy into AI systems. Our recent successful replication of mirror-neuron-like patterns in transformer architectures - demonstrated in follow-up experiments - indicates these experiments may scale to large language models, where intrinsic alignment mechanisms could anchor externally-derived constraints. It advances both our scientific understanding of emergent cognition and the practical pursuit of robust AI alignment, with significant implications for addressing the challenges posed by increasingly capable and autonomous AI.

\section*{Statements and Declarations}

\textbf{Competing Interests:} The author declares no competing interests.  

\textbf{Funding:} This work received no external funding.  

\textbf{Ethics Statement:} This research did not involve human participants or animals.  

\textbf{Author Contributions:} The author conceptualized the study, designed and implemented experiments, analyzed the data, and wrote the manuscript.  

\textbf{Data Availability:} The datasets and code that support the findings of this study are available from the corresponding author upon reasonable request.


\bibliography{references}

\end{document}